\documentclass{svproc}
\usepackage[pdftex]{graphicx}
\DeclareGraphicsExtensions{.pdf,.jpeg,.png}
\usepackage{url}

\usepackage{xcolor}
\usepackage{amsmath,xparse}
\usepackage{hyperref}
\hypersetup{colorlinks,allcolors=red}

\begin{document}
\mainmatter              
\title{Reproducible  Pruning  System  on  Dynamic  Natural  Plants  for  Field Agricultural  Robots}
\titlerunning{Field Agricultural Robotics}  
%
\author{Sunny Katyara\inst{1,2} \and Fanny Ficuciello\inst{2} \and
Darwin G. Caldwell\inst{1} \and Fei Chen\inst{1} \and Bruno Siciliano\inst{2}}
\authorrunning{Sunny Katyara et al.} 
%
\tocauthor{Sunny Katyara, Fanny Ficuciello, Darwin G Caldwell, Fei Chen, Bruno Siciliano}
\institute{Department of Advanced Robotics, Istituto Italiano di Tecnologia, Via Morego 30, 16163, Genova, Italy\\
\and
Interdepartmental Center for Advances in Robotic Surgery, University of Naples Federico II, Naples 80125, Italy\\
}
\maketitle{This research is supported in part by the project ``Grape Vine Perception and Winter Pruning Automation'' funded by joint lab of Istituto Italiano di Tecnologia and Università Cattolica del Sacro Cuore, and the project ``Improving Reproducibility in Learning Physical Manipulation Skills with Simulators Using Realistic Variations'' funded by EU H2020 ERA-Net Chist-Era program}

\begin{abstract}
Pruning is the art of cutting unwanted and unhealthy plant branches and is one of the difficult tasks in the field robotics. It becomes even more complex when the plant branches are moving. Moreover, the reproducibility of robot pruning skills is another challenge to deal with due to the heterogeneous nature of vines in the vineyard. This research proposes a multi-modal framework to deal with the dynamic vines with the aim of sim2real skill transfer. The 3D models of vines are constructed in blender engine and rendered in simulated environment as a need for training the network. The Natural Admittance Controller (NAC) is applied to deal with the dynamics of vines. It uses force feedback and compensates the friction effects while maintaining the passivity of system. The faster R-CNN trained on 3D vine models, is used to detect the spurs and then the statistical pattern recognition algorithm using K-means clustering is applied to find the effective pruning points. The proposed framework is tested in simulated and real environments. 
\end{abstract}
\section{Introduction}
\begin{figure}[t]
      \centering
      \includegraphics[width=12cm]{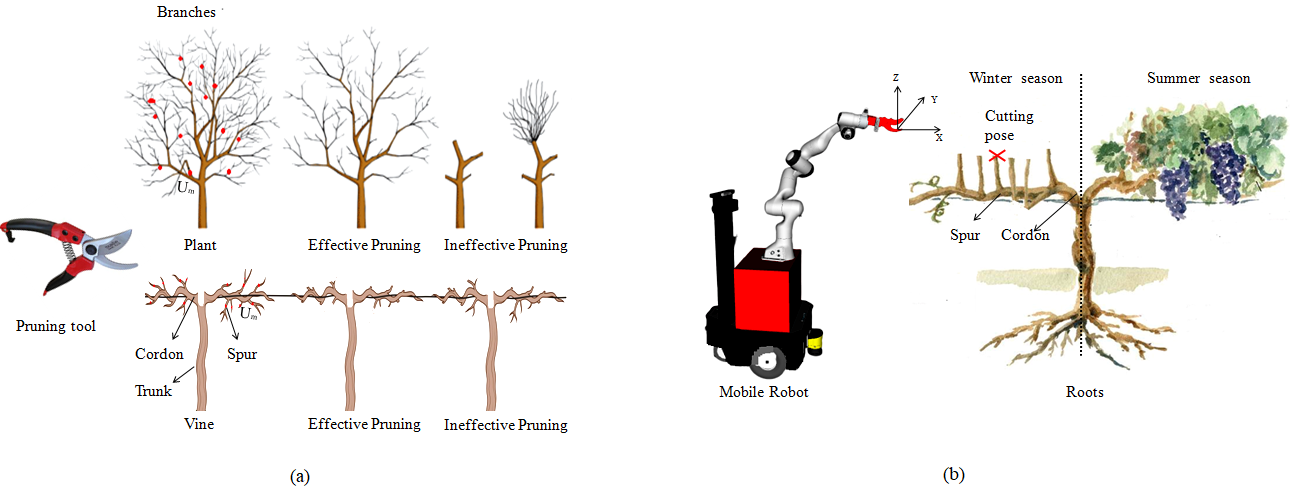}
      \caption{Pruning schema (a) represents general pruning scenarios (b) illustrates the robot pruning system.}
      \label{fig_general_pruning_scenarios}
      \vspace{-15pt}
   \end{figure}

Field robotics is an important research domain especially in agricultural applications as promising solution to increase the yield and reduce the environmental effects. Field robotics uses advanced technologies to give more autonomy to mobile robots to help or do the tasks that are hazardous and dangerous to humans. Now-a-days, most of the agricultural robots are primarily being used for monitoring and quality control such as aerial inspection but can also be used for heavy tasks like harvesting, seeding, \textbf{pruning} etc. As an emerging research domain for agricultural robotics, robotic pruning system is receiving much attention in the community these days. 
Pruning is an important agricultural activity of removing the branches, limbs and spurs from the natural plants. Pruning becomes effective if it only removes the unwanted, unhealthy and poorly positioned branches with least effects on the healthy parts. The effective pruning makes the difference to the quality and yield of plants in the field \cite{c1}. Pruning which requires qualified and experienced professionals, can eventually be performed by intelligent and autonomous robotic system. The stand-alone mobile robot equipped with vision and machine learning is used to travel through the agricultural field and to identify, manipulate and finally cut the branches. The general idea of pruning is illustrated in Fig. \ref{fig_general_pruning_scenarios}. As a representative application for commercial plants, the \textit{grape vine winter pruning} is chosen for analysis. It can be seen from Fig \ref{fig_general_pruning_scenarios}{a} that grape vine pruning is more difficult as compared to other plants, because of complexity in manipulating, locating and deciding the potential cutting points.

Manipulation skills of robot require active perception and interactive learning technologies. Active perception defines the behaviour of agents i.e, robots in the environment on the basis of collected sensor data in order to increase its autonomy and performance. However, robot learning is an iterative process and thus the robot learner tries eventually to learn the specific set of skills and generalize them to new conditions \cite{c2}. Since, in agricultural applications, the reproducibility of manipulations skills is another important aspect because most of the plants even in the same field have different physical characteristics. Therefore, a simulator with real time rendering allows developing large data base and testing the learned skills under the new conditions. Moreover, the \textbf{sim2real} interface allows teaching and interacting with the robots in their virtual environments (simulated) and enables them to learn different manipulation skills in the multi-structured scenarios. In order to create the virtual environments for robots to learn, the \textbf{3D modeling of objects} is one of the key steps towards the object manipulation.

However, manipulation in field robotics has not yet achieved maturity because of uncertainties associated with objects and environments. Most of the machine learning techniques consider fully known, discrete, stationary, deterministic and discrete environments for robot manipulation skill learning and transfer \cite{c3}. But in order to deal with dynamic, unknown, stochastic and continuous environments, more advanced learning approach with promising long-term solution has to be studied, e.g.,  Reinforcement Learning (RL) \cite{c4}. Due to the number of issues such as natural environmental constraints, limited data sets, need for repetition of experiments, limited resources etc the sim2real approach with highly efficient rendering capabilities are chosen to teach the specific set of skills to robots in simulated environments. 

A lot of research has already been done on manipulating rigid body objects but the \textbf{manipulation of soft and moving objects} is still an open question to answer \cite{c5}\cite{c6}. There are many hindrances in automating the manipulation of soft objects and the most common among them is the estimation of their deformation properties because even they are not same among the objects of similar groups. Moreover, in order to deal with moving objects, especially plants, the interaction controllers are needed \cite{c7}. The most popular interaction controllers are impedance and admittance controllers, which offer gentle and fast response. Impedance controller under the large variations of system parameters such as compliance, stiffness, damping and inertia affects the passivity and stability of system \cite{c8}. However, the admittance controller, more precisely Natural Admittance Controller (NAC) proposed by \cite{c9} uses the force feedback to overcome the effects of friction during the interaction without affecting the passivity of system. In our case, NAC is one of the feasible solution to deal with moving plants whose dynamics are unknown during the interaction and thus requires the passivity of system to be maintained. The complex task of pruning has several steps to be followed but the two most important among them are; the perception of agricultural field and identification of spurs on the plants to be pruned and the second is related to motion planning of robot, with the intention of making cuts.

To the best of our knowledge, most of work on robotic pruning has been done on stationary branches and with least consideration on reproducibility of robotic pruning skills. The contributions of this research are; \textbf{(1)} proposing multi-modular framework which combines deep learning (Faster R-CNN) , computer vision (graph morphometry) and robot control (natural admittance controller) for dynamic and reproducible pruning skills \textbf{(2)} Designing 3D models of the grape vines similar to real ones in blender for training the network in the simulator \textbf{(3)} Teaching and testing pruning skills on mobile manipulator in simulated environment initially and then evaluate them on real system. For the sake of convenience, the grape vine winter pruning is taken as an representative case study, shown in Fig. \ref{fig_general_pruning_scenarios}(b).

\section{Related Works}
A systematic literature review on the challenges and solutions for effective pruning, sensing and automation technologies is investigated in \cite{c10}. It points out that the effective pruning relies on the set of key technologies: pruning methods, 3D modelling of dynamic branches, training of robot perception, and handling the dynamics of process. Even though the study is limited to apple tree pruning but its discussion inspires for general natural plant pruning. Another design of robot system for harvesting apple orchard is presented in \cite{c11}. This system uses a redundant robot manipulator equipped with global camera, which can effectively harvest the desired area at a least error rate using an open loop controller. However, the reliability of system need to be improved with grasping status force feedback. A robot system is introduced in \cite{c12} to address the grape vine pruning. The robot system is equipped with multiple cameras for online 3D reconstruction of the vines. Having this geometrical information of vines, the robot plans a collision free trajectory to apply reasonable cuts. However, the research does not take into account the dynamics of vines and their setup uses a tractor pulling a house with robot system inside the vineyard, limiting its wide adoption to different complex conditions. A dual-arm mobile manipulator for pruning and fruit picking is discussed in \cite{c13}. It deals with the non-holonomic constraint introduced by the wheeled platform using recursive Gibbs-Appell method. The system is tested under symmetrical and unsymmetrical conditions during the pruning and fruit picking. The system considered the dynamics of rigid objects but is not tested on moving and soft object manipulation. The effectiveness of algorithm is measured in terms of simplicity, computational timing and repeatability. The idea of sim2real skill transfer for the simple manipulation tasks i.e, object pushing is suggested in \cite{c14}. The idea is based on randomizing the dynamics of simulator to teach the recurrent policies to robot so that it can adapt to different environments in real world. However, this approach can be extended to the manipulation of complex tasks i.e, pruning with the incorporation of deep learning and vision system.     

   \begin{figure}[t]
      \centering
      \includegraphics[width=8cm]{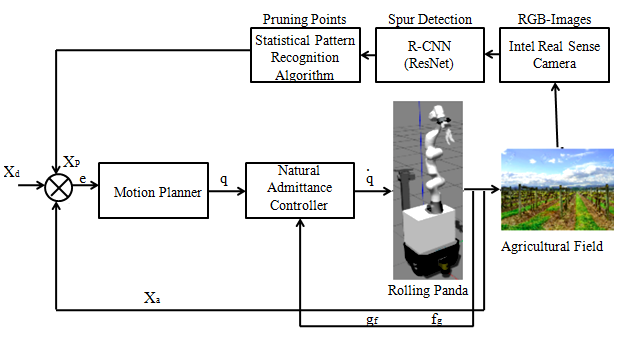}
      \caption{Block diagram of proposed methodology.}
      \label{fig_robot_prunig_chart}
   \end{figure}
   
\section{Research Methodology}

\begin{figure}[t]
      \centering
      \includegraphics[width=8cm]{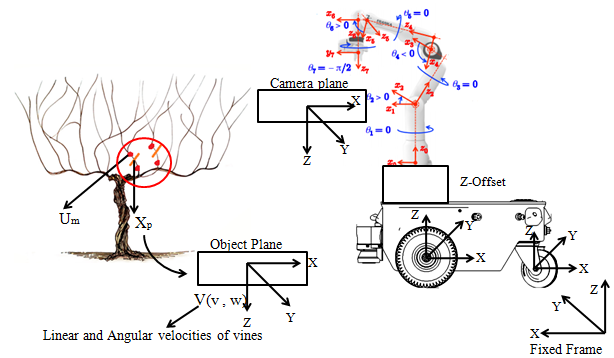}
      \caption{Kinematic model of robot pruning system.}
      \label{fig_robot_prunig_kinematics}
   \end{figure}

For the two important sub-tasks discussed in previous section, initially a convolutional neural network (ResNet-101 R-CNN) trained on modeled 3D vines, is used to detect the possible spurs on the analyzed branches from the set of RGB images and then the graph morphometry together with K-means is applied to find the Potential Pruning Points (PPP) on the vines. Once the pruning points are determined, the next task is to command the robot towards the respective branches on the vines with desired trajectories and hence the quadratic programming is being used for such path planning \cite{c15}. The block diagram of proposed methodology is shown in Fig. \ref{fig_robot_prunig_chart} and associated kinematic model in Fig. \ref{fig_robot_prunig_kinematics}.

\subsection{Modeling of 3D vines and robot platform}

\begin{figure*}[thpb]
      \centering
      \includegraphics[width=11cm]{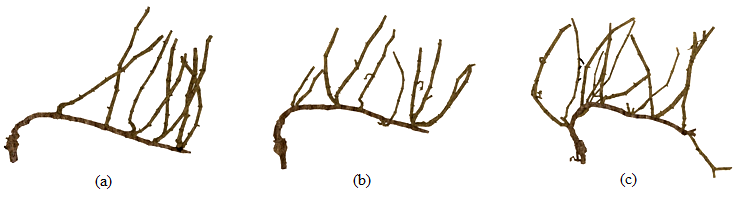}
      \caption{3D models of three different grape vines, (a) illustrates the simpler vine with five spurs, (b) is the vine with six spurs and complex geometry, (c) represents the more complex vine with interconnected shoots on several individual spurs.}
      \label{fig_dynamic_robot_pruning_scenario}
   \end{figure*}

The 3D models of vines are constructed in blender engine and then rendered into simulated environment. The 2D images of vines captured from agricultural field are used as reference for creating 3D models. The 2D images are imported into blender as background in vector graphics form and resized upto 76.3 cm to give them more realistic shape. In the edit mode, the circular extruder with 3.121 cm diameter is selected and the path is traced along the 2D image but with precise rotation, resizing and extension to take into account all the spurs, canes and buds on the vine branches. Once the extrude is done, the proportional editor is activated to give the tilt to the branches according to their real shape. Moreover, the sub-division surface modifier is applied to shape the irregularities on the vines. In order to give real looking texture, UV mapping is done with unwrapping on the selected texture image to follow the lines on real vines and cubic projection is applied to cover the whole surface area of designed vine. Since, the vines are given multi-textures similar to realistic ones and the buds are carefully being placed on alternatives sides of the canes. For adding the softness to vines, specially the jiggle effect when wind blows, the dynamics of vines are changed from rigid body to cloth physics. The pinning is activated and the weighted paint is selected to add jiggle to the branches and then a vertex group is created and scene is activated. In the add brush setting, the paint is applied on the parts in wire frame mode which are not needed to jiggle until they turn reddish. The final result for all the chosen three vines, designed and rendered in blender are shown in Fig \ref{fig_dynamic_robot_pruning_scenario} (a, b, c).  

In this study, the mobile manipulator, named ``Rolling Panda'', is consist of 7-DOF robot arm from \textbf{Franka Emika}, a two wheel mobile platform from \textbf{Neobotix}, two Intel Realse D435i RGBD cameras attached to the end-effector, a cutting tool with pruning shear attached to the end-effector, and a SICK laser scanner for 2D SLAM.

\subsection{Estimation of Potential Pruning Points (PPP)}
 
\begin{figure}[thbp]
      \centering
      \includegraphics[width=8.0 cm]{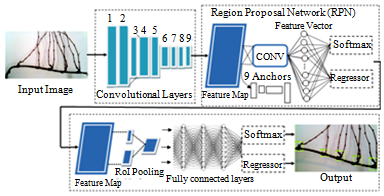}
      \caption{Architecture of faster R-CNN for spur detection.}
      \label{faster rcnn}
   \end{figure}

Before estimating the pruning points, it is required to find the dedicated spur regions so as to narrow down the search space for pruning algorithm. Identification of possible spurs on the vines is the crucial step in finding the pruning region. The Faster R-CNN is used to mathematically characterize the spurs \cite{c16}. It calculates the centroid of each region and assigns bounding box around to ensure the average variance does not go beyond the centroid boundaries. Such information about the centroid of spur is used to find its pose. R-CNN model with ResNet-101 maps the spurs on the vines with mean precision of 0.679 and at a learning rate of 0.0005. The architecture of faster R-CNN network used for spur detection is shown in Fig. \ref{faster rcnn}.

The faster R-CNN uses convolution layers to extract the desired feature map and the anchors are used to define the possible regions from the feature map. Total 9 anchors are used with three aspect ratios 1:1, 1:2 and 2:1 which stride over the feature map skipping 16 pixels at each time. The possible regions predicted by anchors are fed to the regional proposal network (RPN). The RPN is a small neural networks that predicts the region of the object of interest. It actually predicts whether there is object (spur) or not and also frames the bounding box around those spurs. In order to predict the labels of anchors, the output from RPN is passed through softmax or regressor activation functions. Since, RPN results in CNN feature maps of different sizes and hence Region of Interest (RoI) pooling is applied to harmonize the given maps to the same size. The faster R-CNN uses original R-CNN network to perform RoI pooling. It takes the feature vector predicted by RPN, normalize it and passes through two fully connected convolution layers with ReLU activation and thus predicts the class and bounding boxes of object. In order to train and test our faster R-CNN network, 1210 and 245 images of modeled grape vines are used respectively. The dimensions of feature map generated are $25{\times}25$ and the size of anchor is 9, therefore there are total 2916 potential anchors. The test time per image is 0.2 secs and 8359 bounding boxes are detected in the regions of interest with a minimum threshold of 0.7 and which are then finally passed to classifier layer as an output.

\begin{figure}[thbp]
      \centering
      \includegraphics[width=10cm]{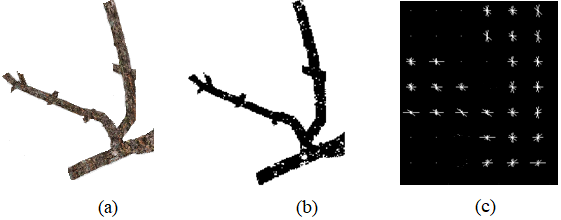}
      \caption{predictive HOG applied to spur region, (a) represents the spur region of interest, (b) illustrates the binary information of region , (c) denotes the predictive histogram of selected spur region.}
      \label{mp_HoG}
   \end{figure}

\begin{figure}[t]
      \centering
      \includegraphics[width=8cm]{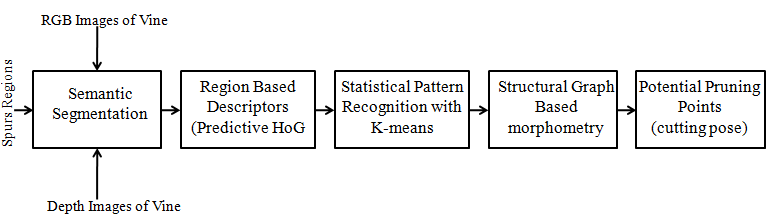}
      \caption{Pipeline for estimation of pruning points.}
      \label{fig_pruning_points}
   \end{figure}

Once the spurs are detected, the next step is to find the Potential Pruning Points (PPP) which mainly depend upon the geometric information of vines and their intrinsic properties. A predictive HOG (Histogram of Gradients) is applied to find out the normalized patches in the spur regions. The regions are subdivided into cells on the basis of gradient intensity and histogram of gradient directions is computed as shown in Fig. \ref{mp_HoG}. The pipeline adopted to estimate the pruning points is defined in Fig . \ref{fig_pruning_points}.

Since, the algorithm results into a flattening matrix with dense information about the intensity gradients. Therefore,the Mean Predictive HOG (\textit{mpHOG}) is calculated under a Bayesian Linear Regression for inference matrix distribution which is normalized by a probabilistic model evidence. On the basis of mpHoG descriptor, a statistical-based pattern recognition pipeline is applied to identify the geometric properties of vines and is merged with K-means clustering to classify the regions in terms of similarity of their intrinsic characteristics. Since, the K-means is a unsupervised machine learning technique used to find the centroid of a given data set and locates the average of data around it \cite{c17}. The algorithm continues to minimize the distance between consecutive centroids until it does not move any more. The algorithm is formulated as minimization problem and is given by Eq. \ref{eqt_O}

\begin{equation}
\bar{F}=\sum_{m=1}^{k}\sum_{n=1}^{j}{||U_m-V_n||^2}
\label{eqt_O}
\end{equation}


The objective function minimizes the distance between $U_m$ points and $V_n$ cluster centre. The structural graph based morphometry uses the graph kernel to identify the similarity between the obtained clusters. The classifier with local information locates the centre of data in each neighborhood, thus provides the information that’s maps the series of pruning points in the given regions. Hence, the cutting pose based on the determined pruning points is obtained by using $\bar{X_p}={R}{t}$. Where R and t denotes the rotation and translation of pruning region with the given set of pruning points. According to Rodrigues rotation formula, with the known unit rotation axis vector r($r_x$ , $r_y$, $r_z$) and the angle of rotation Ø, the R is expressed by Eq. \ref{eqt_R}

\begin{equation}
    R(\phi,r)={(cos{\phi})}{I}+{(1-cos{\phi})}{rr}^{t}+{sin{\phi}}{[r]_{x}}
\label{eqt_R}
\end{equation}

I is the identity matrix of $3 \times 3$ and t is the skew symmetric matrix. Let $p_1$ and $p_2$ be the two pruning points in the region, they should satisfy the constraint defined by the Eq. \ref{eqt_C}

\begin{equation}
    {p_{2}}^{T}{X_{p}}{p_{1}}=0
\label{eqt_C}
\end{equation}

The homogeneous vectors of the pruning points $p_1$ and $p_2$ are always being pre-multiplied by the inverse of intrinsic camera matrix Mc.

\subsection{Motion Planner}

Motion planners are used to navigate the robot to a goal through the sequence of valid configurations. The configurations of robot subjected to constraints are used to avoid collisions with the obstacles and maximize the manipulability of system. The motion planner formulates the task as a cost function to be minimized on the basis of constraints solved by using the equality and inequality solvers and is expressed by the Eq. \ref{eqt_D}. 

\begin{equation}
    {{X}\otimes{Y}}= \{ax-y \in {R}^{n}|x \in X, y \in Y, a \in R\}
\label{eqt_D}
\end{equation}

Where X, Y and R represent goals, constraints and robot configuration space respectively. In order to define the pruning task and to set the priorities on it, a stack is formed. In our research, the Stack of Task (SoT) is formulated by the Eq. \ref{eqt_E}\cite{c18}, with $L_{jointposition},$ and $L_{jointvelocity}$ representing constraints on robot system joint positions and velocities respectively

\begin{equation}
\begin{bmatrix} \{T_{base}\}^{World} \\ \{T_{Arm}\}^{Base}+\{T_{Endeffector}\}^{Arm}  \\ \{T_{Postural}\}^{world} 
\end{bmatrix}
\label{eqt_E}
\end{equation}
\quad \quad \quad \quad \quad \quad \quad \quad \quad \quad \quad \quad \quad \quad \quad \quad s.t 

 \quad \quad \quad \quad \quad \quad \quad \quad \quad \quad $L_{jointposition}+L_{jointvelocity}$

T represents task and is generally defined by ${T_i}=({J_i}, {\dot{e_i}})$

\begin{footnotesize}
\noindent $J_i$ - Jacobian of associated links:\\
\noindent $\dot{e_i}$ - Error in joint velocities
\end{footnotesize}

The tasks are given hard and soft priorities and the constraints are defined by the quadratic programming. The task has hard priority if it does not deteriorate the performance execution of first task while the soft priorities are defined at same task levels and they may influence each other on the basis of their associated weights. In Eq. \ref{eqt_E}, the postural task is set at minimum priority as compared to base, arm and end-effector levels. The tasks related to the arm and end-effector are being solved without affecting the configuration of the mobile base. 

In order to deal with such kind of motion planning, the motion planner using Quadratic Programming (QP) optimization interface is used\cite{c19}. The general task using quadratic programming in joint space is described by the Eq. \ref{eqt_G} and is subjected to $R\dot{q}\leq {C}$ constraints

\begin{equation}
    {\dot{q_i}}=argmin({J_i}{q_i}, {\dot{e_i}})
\label{eqt_G}
\end{equation}

\subsection{Natural Admittance Controller and dynamics of vines}

When dealing with vibrating and moving objects, the friction during the interaction is always ambiguous and passivity of the system is also lost. Natural Admittance Controller (NAC) is the method to analyze the system under the force feedback that results in the loss of passivity due to reduction of inertia during the interaction \cite{c20}. Since, NAC declares the designated values of inertias close to the actual physical values and thus ensures to maintain the passivity by eliminating the friction effects. From the differential kinematics, the relationship between the joint and operational space velocities is given by ${\dot{X_i}}={J_{robot}}{\dot{q_i}}$. Where $\dot{X_i}$ is $7\times1$ vector of velocities of franka arm in operational space and $\dot{q_i}$ are its associated joint velocities and $J_{robot}$ is the Jacobian of franka arm.

The forced applied by the end-effector on the plants at the goal position is determined by Eq. \ref{eqt_I}

\begin{equation}
    {F_e}={K}({P_g}-{P_{base}})-{B}{\dot {X_i}}
\label{eqt_I}
\end{equation}

Where, K and B define the stiffness and damping of panda arm respectively, $P_g$ and $P_{base}$ are the goal points and the base points respectively. The total interaction force is the sum of end-effector force and feedback force from the F/T sensor for plants ($F_{net} = F_{e} + F_{sensor}$). The equivalent joint torque thus is computed by ${\tau_i}={J_{robot}} {F_{net}}$, with obvious meaning of variables. The values of joint accelerations needed to reach the goal are given by Eq. \ref{eqt_K}

\begin{equation}
    {\ddot{q}_{desired}}={H}^{-1}{\tau_i}-{B_H}{\dot{q}_{actual}}
\label{eqt_K}
\end{equation}

Where, H is the matrix of inertias and $B_H$ is the desired damping. With the use of Euler’s integration rule, the joint velocities are computed.

With the control loop running at $100 Hz$, the desired joint velocities are sent to command the robot to interact with stationary and moving vines. The filter is used at force-feedback to suppress the noisy data and also the anti-wind up criteria is applied to avoid the saturation of controller and system parameters. 

The motions of the vines are modeled with three virtual joints as PRR kinematic chain in simulated environment, which can translate in and rotate around $y$ direction and also generates rotation around $z$-axis. The dynamics of plant is defined by ${F_{plant}}={m}{\dot{v}}+{b}{v}$

Where, ${F_{plant}}$ is the force applied by plant and its sensed by F/T sensor as ${F_{sensor}}$, m represents the mass of plant, $\dot{v}$ is its acceleration and b is the friction. The desired admittance of the system in Laplace domain is defined by Eq. \ref{eqt_M}

\begin{equation}
    {C_{desired}(S)}=\frac{1}{{m}{s}+{B}+{K/s}}
\label{eqt_M}
\end{equation}

Since, the mass of vine is not altered but the friction b is compensated to get the desired behavior with the feedback gain given by Eq. \ref{eqt_N} and the effect of friction further diminishes with an increase of velocity loop gain $g_v$. It has been found that the system under NAC is passive, no matter how large the value of $g_v$ is increased for minimizing the friction effects. 

\begin{equation}
{g_f}=\frac{({b}-{g_v}-{B}){s}-{K}}{{m}{s}^{2}+{B}{s}+{K}}
\label{eqt_N}
\end{equation}

\section{Results And Discussion}

   \begin{figure}[h]
      \centering
      \includegraphics[width=11cm]{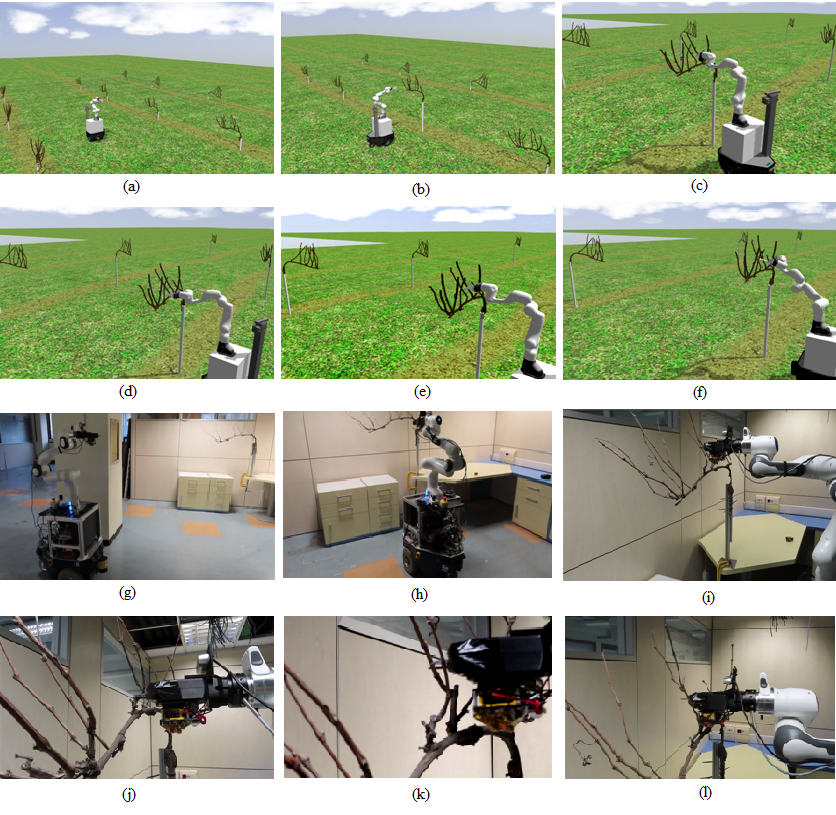}
      \caption{Robot pruning scenarios in simulated and real environments, (a and g) represent the mobile manipulator navigating the field, (b and h) illustrate the pose of mobile manipulator reaching the designated vine, (c and i), (d and j), (e and k), (f and l) report the first, second, third and forth cutting actions respectively.}
      \label{pruning system scenarios}
   \end{figure} 
   
\begin{figure}[h]
      \centering
      \includegraphics[width=12cm]{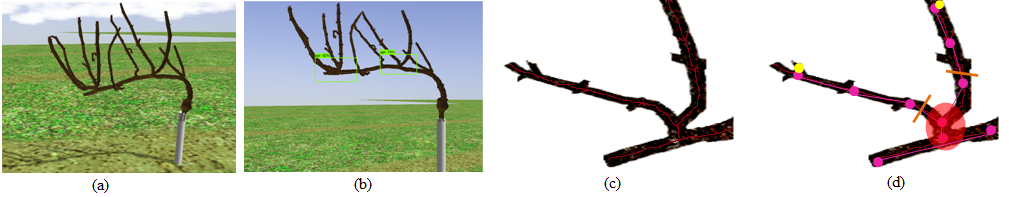}
      \caption{Detection of spurs and pruning points on vines. in (a) the chosen vine for pruning is shown, in (b) the possible spurs on the selected vine are found, in (c) the geometrical data of vine within the spur region is shown (d) the location of pruning points, region and cutting pose are illustrated.}
      \label{fig_spurs_pruning}
   \end{figure}

The results from the spur detection algorithm trained for almost 6 hours are shown in Fig. \ref{fig_spurs_pruning}(a, b). The bounding boxes represent the possible spurs on the vines with confidence interval of more than $70\%$ in Fig. \ref{fig_spurs_pruning}(b). Figures \ref{fig_spurs_pruning}(c, d) illustrate the Potential Pruning Points (PPP) obtained with the application of proposed algorithm. Fig \ref{fig_spurs_pruning}(c) represents the morphometry data of vine. The big circle in Fig \ref{fig_spurs_pruning}(d) on the regions indicates the classification of spurs, either to prune or not and yellow dots in each region indicate the range of pruning points detected. The pink dots denote the effective PPP on the vine to perform the required cuts.

The cutting pose shown by orange line in the Fig. \ref{fig_spurs_pruning}(d) between pruning points is sent to the motion planner to generate the desired trajectories for the robot platform, executed within 60 secs approximately. The application of proposed framework on robot pruning scenario in simulated and real environments is reported in Fig. \ref{pruning system scenarios}. It can be seen from the Fig. \ref{pruning system scenarios} that the results in the simulated environment Fig. \ref{pruning system scenarios}(a-f) are successfully reproduced on the real robot Fig. \ref{pruning system scenarios}(g-l) with minor deviations in pruning time and pruning accuracy. Total two spurs, 8 potential pruning points and four cutting poses are detected within 6 secs in both the environments successfully. Since, during the experiments on the real robots it is found that it skips one cut on the designated vine position while the robot in simulated environment performs all the four cuts more accurately and in a less amount of time. It is obvious, because the changing real time conditions i.e, lighting, temperature, sensor fidelity i.e, F/T information, friction of vines etc are however steady in the simulated vineyard scenario. However, the domain randomization plays an important role in bridging the gap between sim2real skill transfer but still the $100\%$ reproduciblity of manipulation skills on real system suffers mainly due to inaccuracy in modeling of system dynamics and varying environmental conditions as what happens in our case too.

\begin{figure}[h]
      \centering
      \includegraphics[width=12cm]{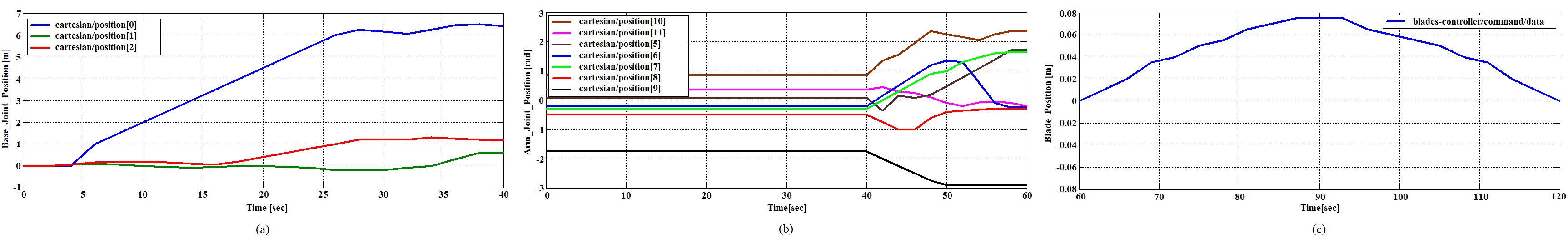}
      \caption{Whole body motion control of robot platform. (a) shows the trajectories followed by base to reach the desired location, in (b) the trajectories to reach selected vine for the franka arm are shown, (c) illustrates the pruning action by cutting blades attached to end-effector.}
      \label{fig_trajectories}
   \end{figure}

The associated joint position trajectories for the base and arm are shown in Fig. \ref{fig_trajectories} (a, b) respectively. It can be seen that the pose of robot arm is not affected when the mobile base is navigated to a desired vine under 40 sec time interval and then the robot arm is actuated to reach the cutting region without violating the current configuration of mobile base. This case validates the definition of hard constraints between the mobile base and robot arm used in our framework.    

\begin{figure}[h]
      \centering
      \includegraphics[width=12cm]{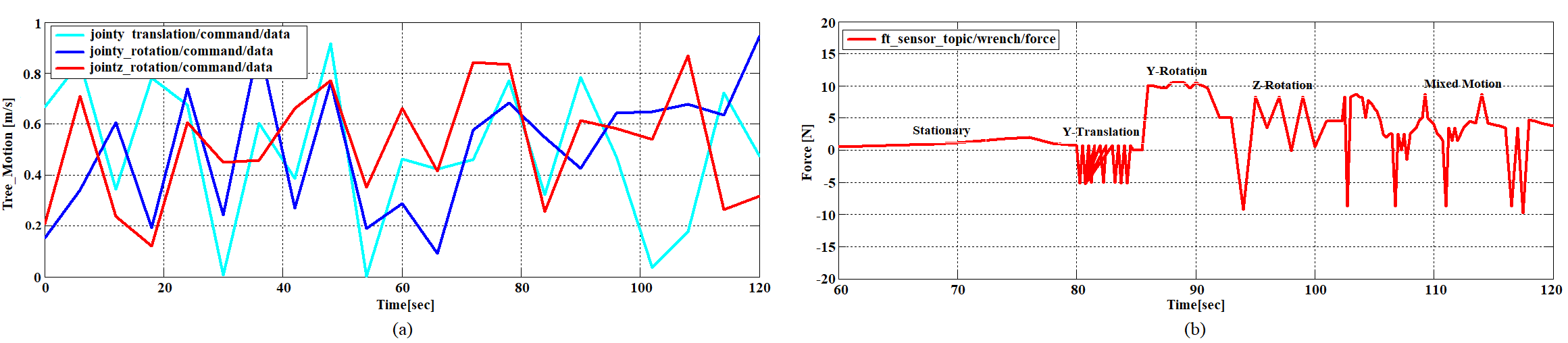}
      \caption{Dynamics of vine branches. (a) denotes designated motions of vines, (b) shows the force profile of vines in static and moving conditions.}
      \label{fig_tree_motion}
   \end{figure}

However, the soft constraints are defined between the cutting tool and robot arm in order to facilitate the adaptive translation and orientation to reach the desired cutting pose on the vines. The vines are initially stationary and the cutting tool after reaching the pose of PPP, performs the required cut and is shown in Fig. \ref{fig_trajectories}(c). The tool centre point (TCP) is defined on the cutting tool at 5 mm away from tip of the blades. It can be seen that the blades on the cutting tool moves $0.075$ m close to each other to execute the pruning action. Next, the vines are given motion in different directions i.e, $y$-translation, $y$-rotation, $z$-rotation and finally combination of all (mixed). The motion of plants within all defined directions is shown in Fig. \ref{fig_tree_motion}(a).

Due to the interaction between the robot arm and vine, the force is exerted by both on each other. The force exerted by end-effector on plant is constant regardless of the motion of vines. The force profile of vine in stationary and dynamic conditions is shown in Fig. \ref{fig_tree_motion}(b). Since the vines are placed in Y-direction and when they touch the cutting tool during the motion equipped with F/T sensor, it experiences a spike of impact as shown in Fig. \ref{fig_tree_motion}(b). The force applied by the end-effector $F_e$ is composed of two terms, one is the force of spring between the arm and vine and the other is due to the damper (inertia) of arm acting between reference and actual velocities of the end-effector.  

In case of dynamic pruning, the F/T information is crucial for effective pruning. When cutting tool and moving vine interact, the force is experienced by F/T sensor and is sent to the planner via NAC as feedback to execute a cut on the vine at the instant of interaction. Since, the cutting action defines the number of attempts made to reach the final assigned end goal. During stationary pruning, the cutting tool mostly performs all the desired cuts in one trial but for moving vines, several attempts are being made to finally reach the required cut.  

The various pruning tests under different scenarios are listed in Table \ref{table_example}. The pruning accuracy is measured in terms of the cuts made by blades at the assigned location with a correct position and orientation \cite{c9}. Since, the value of vine force is averaged over its time to reach the pruning points. The more complex pruning scenario is obviously a mixed one, in which the vines have given random motions in all the defined directions and hence the cutting tool after sensing force tries several times to reach the final cut. 

\begin{table}[t]
\caption{Different Setups of Pruning Scenarios.}
\label{table_example}
\begin{center}
\begin{tabular}{p{1cm}|p{1.5cm}|p{5em}|p{2cm}|p{2cm}|p{2cm}}
\hline
\textbf{Sr. No.} & \textbf{\hfil Motion} & \textbf{Vine Force (N)} & \textbf{Pruning time (sec)} & \textbf{Pruning accuracy (\%)} & \textbf{Pruning actions (times)} \\
\hline
\textit{\hfil1} & {Stationary} & {\hfil2.31494} & {\hfil18} & {\hfil100} & {\hfil1} \\ 
\hline
\textit{\hfil2} & {\hfil Y-trans} & {\hfil6.12634} & {\hfil31} & {\hfil97.2} & {\hfil3} \\
\hline
\textit{\hfil3} & {\hfil Y-rot} & {\hfil9.79863} & {\hfil24} & {\hfil98.7} & {\hfil2} \\
\hline
\textit{\hfil4} & {\hfil Z-rot} & {\hfil10.02314} & {\hfil28} & {\hfil96.4} & {\hfil3} \\
\hline
\textit{\hfil5} & {\hfil Mixed} & {\hfil19.36213} & {\hfil49} & {\hfil95.1} & {\hfil5} \\
\hline
\end{tabular}
\end{center}
\end{table}

\section{CONCLUSIONS}

This research proposed a multi-modal framework for automating the pruning of dynamic vines. For the reproducibility of pruning skills under different scenarios with the aim of sim2real skill transfer, the 3D models of vines were constructed and rendered in blender. The RGB images taken with Intel Real-Sense camera mounted on end-effector were used for spurs detection using Faster RCNN at a threshold of 0.7. The robot arm then moved close to a bounding box at the distance of $8.35cm$ between the cutting tool and vine and then within the vicinity of spurs, the pruning points were located using statistical pattern recognition algorithm based on the graph morphometry. The cutting poses of these points were constructed to command the robot. The vines were initially stationary and then moved in random directions. The proposed framework was trained and tested on designed 3D models of vines in simulators and then the pruning skills were also repeated on the real system in laboratory environment. NAC was applied to deal with dynamic nature of the vines. It was found that the pruning timing was compromised while dealing with moving vines but the accuracy was still maintained within the $\pm 5\%$ tolerance limits.

The current research has been carried out in the controlled laboratory environment with limited sim2real skill transfer. The next goal is to test the learned skills in real vineyard on the field and also to increase the autonomy of robot platform by using deep SLAM under distinct environmental conditions.

\end{document}